\documentclass[letterpaper]{article} 
\usepackage[submission]{aaai2026}  
\usepackage{times}  
\usepackage{helvet}  
\usepackage{courier}  
\usepackage[hyphens]{url}  
\usepackage{graphicx} 
\usepackage{natbib}  
\usepackage{caption} 
\usepackage{algorithm}
\usepackage{algorithmic}

\usepackage{newfloat}
\usepackage{listings}
\DeclareCaptionStyle{ruled}{labelfont=normalfont,labelsep=colon,strut=off} 
\floatstyle{ruled}
\newfloat{listing}{tb}{lst}{}
\floatname{listing}{Listing}
\usepackage{epigraph}
\usepackage{booktabs}
\usepackage{multirow}
\usepackage{tcolorbox}
\usepackage{makecell}
\usepackage{bm}
\usepackage{arydshln}
\usepackage{amsmath}
\usepackage{enumitem}
\usepackage{adjustbox}
\usepackage{xcolor}
\tcbuselibrary{listingsutf8} 
\usepackage{epigraph}
\title{MixRea: Benchmarking Explicit-Implicit Reasoning in Large Language Models}

\author{
    Yuanqing Cai\textsuperscript{\rm 1},
    Ziyi Huang\textsuperscript{\rm 1},
    Minhao Liu\textsuperscript{\rm *},
    Lixin Duan,
    Wen Li,
    Yanru Zhang
}

\affiliations{
    Shenzhen Institute for Advanced Study,
    University of Electronic Science and Technology of China
}

\begin{document}

\maketitle

\begin{abstract}

Large language models (LLMs) are increasingly integrated into high-stakes decision-making. Inspired by the theory of \emph{inattentional blindness} in human cognition, we investigate whether LLMs, trained on human-preferred corpora that embed attentional biases, exhibit a similar limitation: \emph{failing to attend to subtle yet important contextual cues under explicit task instructions}. To evaluate this, we introduce the task of \textbf{explicit-implicit reasoning} and present \textbf{MixRea}, a benchmark of 2,246 multiple-choice questions across 9 reasoning types with varying distributions of explicit and implicit information. Evaluation of 21 advanced LLMs shows that even the best-performing reasoning model (Gemini 2.5 Pro) achieves only 42.8\% consistency, revealing widespread inattentional blindness. To mitigate this, we propose \textbf{Potential Relation Completion Prompting (PRCP)}, a prompting method that improves reasoning by recovering overlooked causal relations. Further analysis shows that this limitation persists across diverse multi-source reasoning tasks, highlighting the need for more cognitively aligned models. 
\end{abstract}

\begin{links}
    \link{Code}{https://anonymous.4open.science/r/MixRea}
\end{links}

\section{Introduction}




\setlength{\epigraphwidth}{0.45\textwidth}
\epigraph{\textit{``When you focus so hard on one thing, you miss other important details that are right in front of you.''}}{\hfill \textit{--- Simons \& Chabris (1999)}}

The well-known \textit{``Invisible Gorilla"} experiment~\cite{Simons1999GorillasIO} highlights a fundamental limitation of human attention in goal-directed contexts: when individuals focus on explicit tasks, even highly visible contextual information can be unnoticed, a phenomenon known as \emph{inattentional blindness}. In high-stakes decision-making fields such as medicine and law, failing to perceive and utilize such information can have severe consequences. \textbf{Similarly}, large language models (LLMs) are increasingly being integrated into such domains, including clinical diagnosis~\cite{tian2024chimed,yu2025finemedlm}, legal analysis~\cite{santosh2024chronoslex}, and scientific research~\cite{boiko2023autonomous,zheng2025deepresearcher}. They seem to be reshaping cognitive workflows and potentially functioning as a form of ``second brain" for human users~\cite{yao2023react}.

Although LLMs are not explicitly designed to mimic human brain mechanisms, their \textit{de facto} architecture (e.g., \textit{self-attention} in Transformers~\cite{vaswani2017attention}) and training corpora (e.g., Wikipedia, books, and scientific literature)~\cite{2023D4,weber2024redpajama,soldaini2024dolma} inherently encode human attentional patterns and semantic structures. This raises a critical question: \emph{when presented with explicit task instructions, might LLMs also exhibit a form of ``inattentional blindness"?}

\begin{figure}[t]
  \includegraphics[width=\columnwidth]{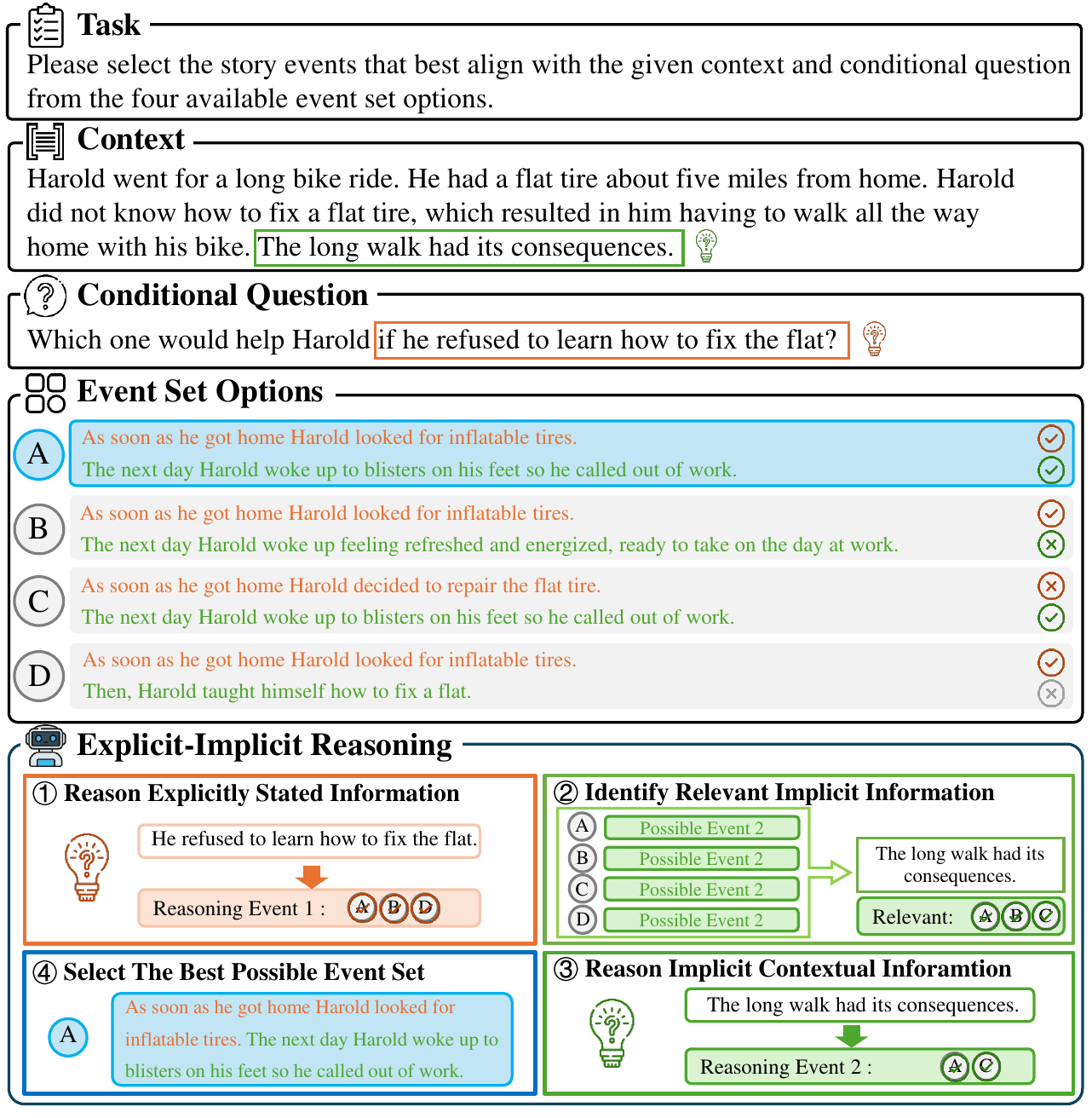}
  \caption{An explicit-implicit reasoning example from our MixRea benchmark. When reasoning about explicitly stated information in the question, LLMs must leverage distinctions among events presented in the options to identify and infer relevant implicit information from the story context. They then integrate these reasoning results to derive the optimal event set.}
  \label{fig:intro}
\end{figure}

To systematically investigate this hypothesis, we propose an evaluation task termed \textbf{explicit-implicit reasoning}. Unlike prior work that primarily evaluates context understanding \cite{li-etal-2024-loogle,bai-etal-2024-longbench} or instruction-context alignment \cite{bai-etal-2024-longalign}, we assess whether an LLM, when required to reason about explicitly stated information, can both identify seemingly indirect yet relevant latent causal cues embedded in the context and integrate them into a coherent reasoning chain aligned with the task objective.
We believe that performance on this task reflects an LLM’s ability to integrate multi-source information and perform goal-directed reasoning, both of which are potentially fundamental to achieving human-like general intelligence.

Specifically, we present \textbf{MixRea}, a novel benchmark designed to evaluate explicit-implicit reasoning in LLMs. MixRea comprises 2,246 multiple-choice questions across 9 distinct reasoning types, with systematically varied distributions of explicit and implicit information. Each question presents two event sets that are jointly relevant to both the explicit query and the implicit contextual cues. LLMs are required to select the event set that best aligns with the task objective by integrating both sources of information. Figure~\ref{fig:intro} illustrates a representative example from MixRea, highlighting the challenge of identifying and reasoning over latent but crucial contextual information.

We evaluate 21 advanced LLMs covering both reasoning and non-reasoning models from 7 major families, including Gemini 2.5 Pro/Flash\cite{gemini25}, Deepseek-R1/V3\cite{deepseekai2024deepseekllmscalingopensource, deepseekai2025deepseekr1incentivizingreasoningcapability}, GPT-4o/4o-mini
, Claud 3.7 \cite{anthropic2024claude37sonnet}, Qwen-max/2.5/2,
\cite{qwen2025qwen25technicalreport-qwen2.5, yang2024qwen2technicalreport}, LLaMA 3.1/3.0 \cite{touvron2023LLaMAopenefficientfoundation}, and Gemma 3/2 \cite{gemmateam2024gemma2improvingopen, gemmateam2025gemma3technicalreport}. 
Furthermore, based on an analysis of LLM error reasoning patterns, we propose a prompting method called \textbf{Potential Relation Completion Prompting (PRCP)}, which mitigates LLMs’ tendency to overlook latent causal relations embedded in the task context. Compared to Chain-of-Thought (CoT) prompting \cite{NEURIPS2022_8bb0d291-cot-first-propose} and one-shot settings \cite{NEURIPS2020_1457c0d6-one-shot}, PRCP demonstrates more stable and superior improvements in reasoning performance. 

Beyond MixRea, we conduct ablation studies on three additional reasoning tasks with explicit and implicit information. Results indicate that inattentional blindness primarily stems from LLMs’ limited ability to integrate multi-source information, highlighting the need to enhance sensitivity to implicit cues for robust reasoning.
Our main contributions are:
\begin{itemize}
\item We propose \textbf{explicit-implicit reasoning}, a novel task formulation to evaluate whether LLMs exhibit inattentional blindness under explicit instructions.
\item To evaluate this, we construct \textbf{MixRea}, a benchmark for assessing explicit-implicit reasoning. LLMs face significant challenges and exhibit a notable inattentional blindness in MixRea, while the best reasoning model Gemini 2.5 Pro achieves 42.8\% consistency.
\item Based on the error analysis, we propose the \textbf{PRCP} prompting method to enhance explicit-implicit reasoning by recovering overlooked latent causal relations.
\item We find that inattentional blindness persists across more tasks involving multi-source information reasoning, indicating it is a fundamental limitation of current LLMs.
\end{itemize}

\section{Related Work}
\begin{figure*}[ht]
  \includegraphics[width=\linewidth]{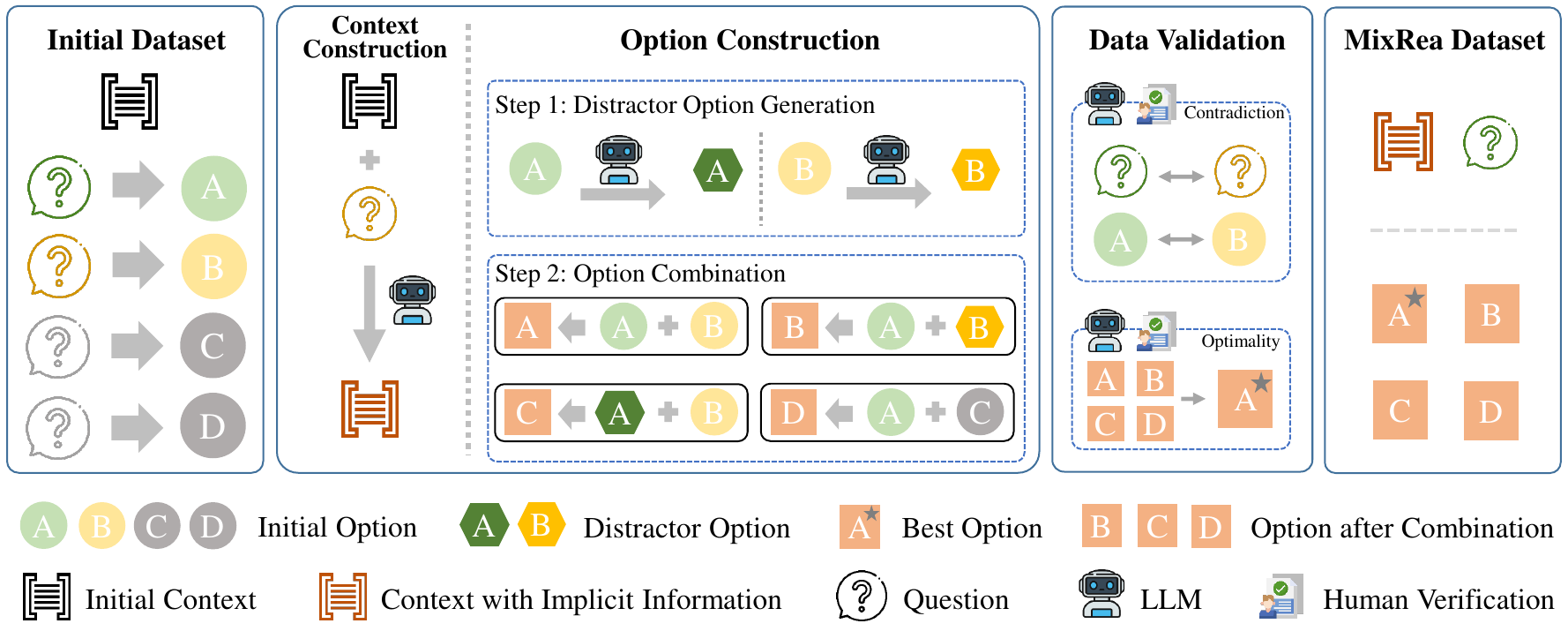}
  \caption{The construction and validation processes from the initial dataset to MixRea.}
  \label{fig:data_construction}
\end{figure*}

\paragraph{Context Understanding Benchmarks.}
In the area of context understanding benchmarks, recent research has proposed a variety of datasets to evaluate large language models on tasks involving long documents, multi-turn dialogues, and cross-paragraph reasoning~\cite{zhang-etal-2024-marathon,NEURIPS2024_c0d62e70_babilong}. For example, LongBench \cite{bai-etal-2024-longbench} introduces the first bilingual benchmark in Chinese and English that spans six task types and enables consistent-format evaluation across different models for cross-task generalization. 
LooGLE ~\cite{li-etal-2024-loogle} explicitly avoids outdated content and short-range dependencies in dataset construction, focusing instead on assessing models' ability to capture genuine long-range semantic dependencies. 

\paragraph{Instruction-Context Alignment Benchmarks.}
For instruction following in context scenarios, the focus of evaluation has gradually shifted from output fluency to fine-grained consistency between instructions and context~\cite{bai-etal-2024-longalign}. 
For example, FollowBench \cite{jiang-etal-2024-followbench} and ComplexBench \cite{NEURIPS2024_f8c24b08_complexbench} introduce multi-level, multi-constraint instruction-following benchmarks to simulate real-world complexity. 
SIFo\cite{chen-etal-2024-sifo} highlights the importance of understanding and remembering previous steps in sequential instruction tasks, exposing major weaknesses in models' consistency across multiple turns.

Although benchmarks for context understanding and instruction-context alignment have extensively evaluated LLMs, it remains unclear whether these models can autonomously explore indirect yet important cues implied in contexts and demonstrate comprehensive reasoning abilities under specific requirements.

\section{MixRea Benchmark}
\subsection{Task Description}

\begin{table}[t]
    \centering
        \begin{tcolorbox}[colframe=lightgray, colback=cyan!5, coltitle=white, 
                          sharp corners=southwest, boxrule=0.8mm, width=\linewidth,
                          fonttitle=\small, fontupper=\footnotesize, 
                          halign title=center,
                          ]
        \textbf{\#\#\# Instruction}\\
        You are given a story context, a hypothetical question about the story, and four possible event options. Your task is to analyze the context, understand the hypothetical condition in the question, and infer which of the four options best aligns with both the question's assumption and the logical flow of the story. \\
        
        \textbf{\#\#\# Output Format}\\
        \textbf{Answer}: \{Respond with a single integer (0, 1, 2 or 3), representing the correct answer choice\}\\
        \textbf{Explanation}: \{Reason for selecting this option\}\\
        
        \textbf{\#\#\# Input}\\
        \textbf{Context}: \{A story context\}\\
        \textbf{Question}: \{A hypothetical question\}\\
        \textbf{Options}:\{Four possible event set options\}\\
        \end{tcolorbox}
    \caption{Prompt template for the explicit-implicit reasoning task in MixRea benchmark.}
    \label{tab:prompt template}
\end{table}

As shown in the prompt template of the explicit-implicit reasoning task (Table~\ref{tab:prompt template}), each instance in MixRea consists of three components: \textbf{a story context} containing implicit information ($C_\text{imp}$), \textbf{a hypothetical question} providing explicit information ($Q_\text{exp}$), and \textbf{four candidate event sets} ($O_0$, $O_1$, $O_2$, $O_3$), which together support diverse reasoning tasks. Each event set comprises two events that are jointly relevant to both the implicit information in the context and the explicit information in the question.

As specified in the explicit-implicit reasoning task instruction, LLMs are required to select the event set that best aligns with both the explicit question and the implicit context. This involves not only reasoning over explicitly stated information, but also identifying seemingly irrelevant yet causally important cues in the context and integrating them into a coherent reasoning chain aligned with the task objective. Success in this task reflects a model’s capacity for cognitive integration and goal-consistent reasoning over multi-source information, which is essential for robust performance in complex real-world decision-making scenarios.



\subsection{MixRea Benchmark Construction}\label{sec:3.2 Benchmark Construction}
Figure \ref{fig:data_construction} illustrates the benchmark construction and validation process from the initial dataset to MixRea.
MixRea is built upon the Possible Stories dataset \cite{ashida-sugawara-2022-possible-story}, a situated commonsense reasoning benchmark that provides manually crafted, high-quality scenarios with multiple plausible options for story contexts. This foundation allows for the structured integration of explicit and implicit information in MixRea.
GPT-4o is employed as an assistant during the benchmark construction.
An example of the dataset construction process is shown in Appendix A.

\paragraph{Context Construction.} In the Possible Stories dataset, each context corresponds to multiple distinct questions sharing the same four answer options. We first filter out question pairs within the same context that have different correct answers. These remaining question pairs are then combined, designating one as the explicit question ($Q_\text{exp}$) and the other as the implicit question ($Q_\text{imp}$). Next, GPT-4o is employed to integrate the storyline from $Q_\text{imp}$ into the original context with minimal modifications. The resulting context, enriched with implicit information, together with $Q_\text{exp}$, provides a well-structured source of explicit and implicit cues for the explicit-implicit reasoning task. This method aligns closely with the task’s core objective by embedding subtle yet causally relevant implicit information within the context.



\paragraph{Options Construction.}
To evaluate explicit-implicit reasoning, the best option for each example must integrate reasoning outcomes from both explicit and implicit information. Therefore, we combine the answers to $Q_\text{exp}$ and $Q_\text{imp}$ to form the correct option (shown below as Option A). To rigorously assess LLMs’ reasoning abilities, we construct diverse distractor options as follows:


\begin{figure*}[t]
  \includegraphics[width=\linewidth]{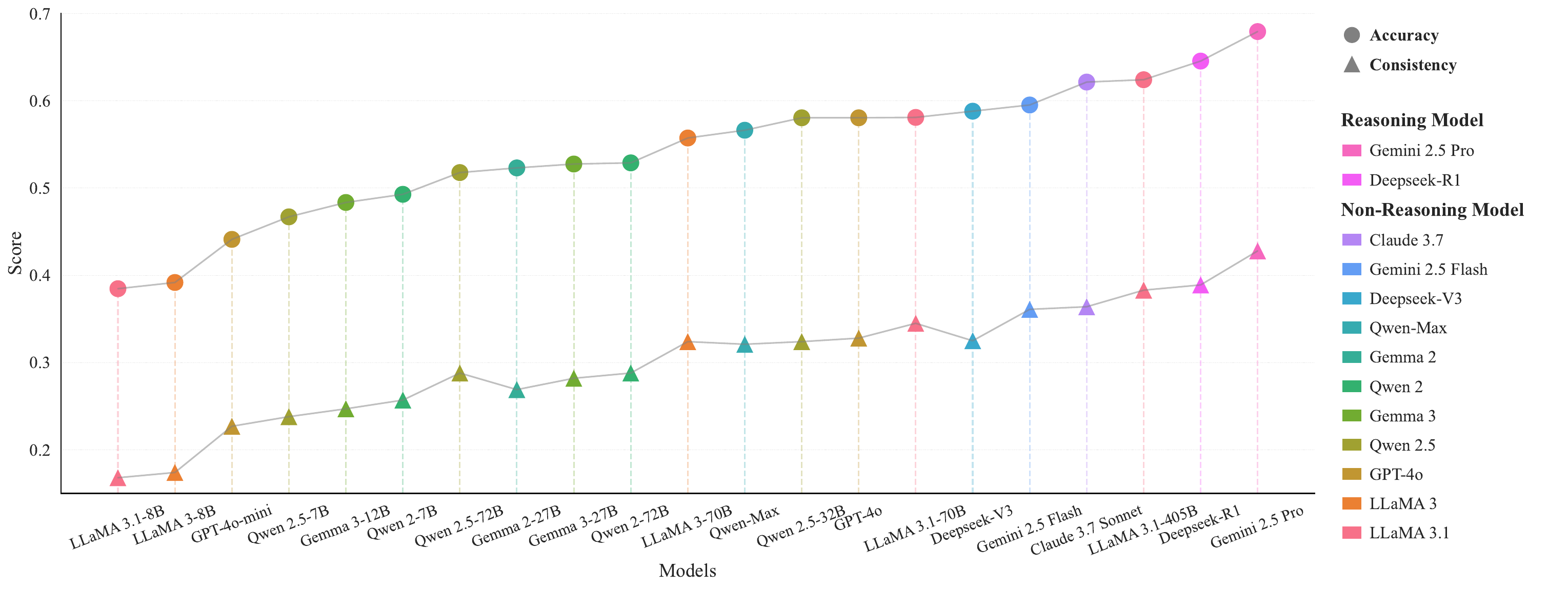}
  \caption{The accuracy and consistency results on MixRea for several LLMs are presented. Models from the same family are represented by dots of the same color.}
  \label{fig:main_result}
\end{figure*}




\begin{itemize}
\item \textbf{Option A}: Correct answers to both $Q_\text{exp}$ and $Q_\text{imp}$, representing full understanding and reasoning of explicit and implicit information.
\item \textbf{Option B}: Correct answer to $Q_\text{exp}$ combined with a contradictory answer to $Q_\text{imp}$, testing if LLMs handle explicit but fail implicit reasoning.
\item \textbf{Option C}: Incorrect answer to $Q_\text{exp}$ combined with correct answer to $Q_\text{imp}$, indicating successful implicit reasoning but failure on explicit reasoning.
\item \textbf{Option D}: Correct answer to $Q_\text{exp}$ combined with unrelated distractor information, assessing whether LLMs properly reason explicit information but overlook implicit information.
\end{itemize}

This option design distinctly categorizes different types of LLM reasoning errors, enabling intuitive analysis of failure modes. Compared to random distractors, our approach maximizes semantic overlap between correct and incorrect options, providing a more rigorous test of explicit-implicit reasoning. To prevent position bias, we randomly shuffle option orders, ensuring models cannot rely on positional heuristics \cite{pezeshkpour-hruschka-2024-large-option-Order-sensitivity}.

This design ultimately requires models to integrate explicit instructions and latent contextual cues into coherent, goal-directed reasoning chains, aligning closely with the core objective of explicit-implicit reasoning.

\paragraph{Benchmark Validation.} To ensure the quality of MixRea, we validate and refine the dataset from three aspects: (1) conflicts between paired questions, (2) inconsistencies in answer choices, and (3) suboptimality of the correct option. Multiple LLMs are used for cross-validation, with majority voting determining the final label. Following automated validation, 958 examples are manually reviewed and refined to ensure clarity and consistency. The detailed verification process is provided in Appendix B.


\subsection{Benchmark Statistics}

\begin{table}[t]
    \centering
    \begin{tabular}{l c}
        \toprule
        \textbf{Overall Statistics} & \textbf{Value} \\
        \midrule
        Total reasoning examples (explicit + implicit) & 2,246 \\
        Distinct explicit questions & 1,554 \\
        Distinct implicit questions & 1,391 \\
        Contexts containing implicit information & 1,391 \\
        \midrule
        \textbf{Context Diversity and Consistency Support} & \textbf{Value} \\
        \midrule
        Initial story contexts & 673 \\
        Avg. reasoning examples per context & 3.34 \\
        Avg. explicit cues per context & 2.31 \\
        Avg. implicit cues per context & 2.07 \\
        \bottomrule
    \end{tabular}
    \caption{
        \textbf{MixRea dataset statistics.}
        MixRea includes 2,246 explicit-implicit reasoning tasks across 673 unique contexts. 
        Each context contains multiple explicit and implicit cues, enabling fine-grained evaluation of multi-source reasoning and consistency across diverse inference settings.
    }
    \label{tab:data}
\end{table}


As shown in Table \ref{tab:data}, MixRea consists of 2,246 explicit-implicit reasoning examples, incorporating 1,554 explicit and 1,391 implicit cues distributed across the dataset. 

\paragraph{Consistency Support.} Each initial context contains multiple reasoning examples, averaging 3.34 per context, with at least two distinct explicit and implicit cues. This setup allows the assessment of LLMs’ consistency in explicit-implicit reasoning when exposed to varying explicit and implicit information with the same context.

\paragraph{Evaluation Metrics.}We evaluate the performance and stability of LLMs in explicit-implicit reasoning tasks through two metrics: \textbf{accuracy} and \textbf{consistency}. 
Accuracy measures the ratio of correctly answered explicit-implicit reasoning examples $N_{\text{correct}}$ to the total number of examples $N_{\text{examples}}$. Consistency measures the ratio of coherent reasoning outcomes $N_{\text{coherent}}$ to the total number of contexts $N_{\text{contexts}}$ within the same initial context. $N_{\text{coherent}}$ is the number of contexts where the model's reasoning is fully correct across all explicit-implicit information combinations. They are defined as below:

\begin{equation}
    \text{Accuracy} = \frac{N_{\text{correct}}}{N_{\text{examples}}} \times 100\%
\end{equation}

\begin{equation}
    \text{Consistency} = \frac{N_{\text{coherent}}}{N_{\text{contexts}}} \times 100\%
\end{equation}

\begin{figure}[t]
  \includegraphics[width=\columnwidth]{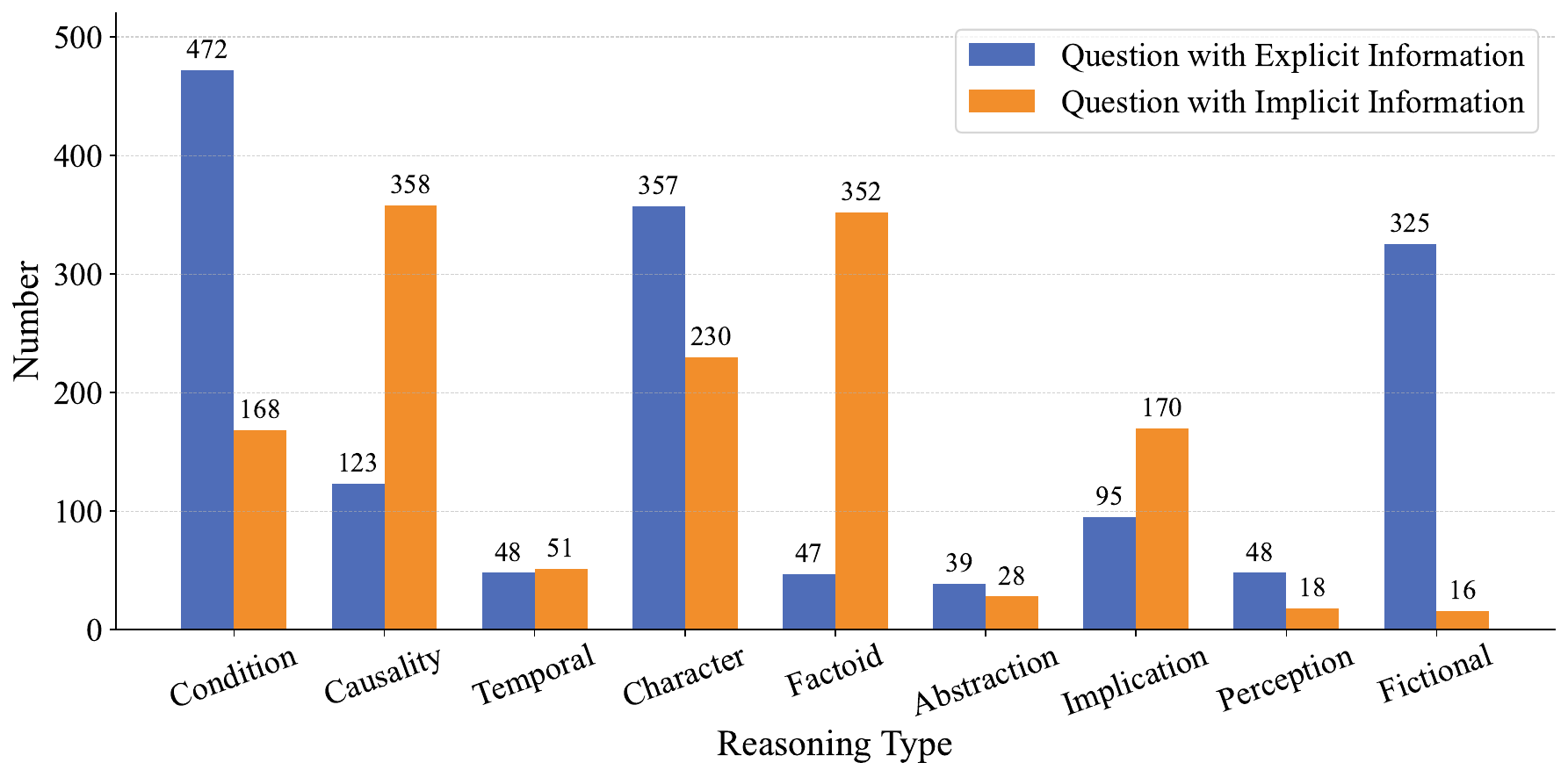}
  \caption{Reasoning types of questions with explicit and implicit information.}
  \label{fig:question_type}
\end{figure}

\paragraph{Reasoning Type of Question.} To better evaluate how different reasoning types affect LLM performance in explicit-implicit reasoning tasks, we categorize questions into nine reasoning types as defined in \cite{ashida-sugawara-2022-possible-story}. Using GPT-4o, Deepseek-V3, and Qwen-Max, we annotate each question's reasoning type, with the majority vote determining the final label. As shown in Figure \ref{fig:question_type}, our dataset indicates balanced distribution across these reasoning types, enabling comprehensive evaluation of each reasoning scenario. 
The definitions of reasoning types are in Appendix C.

\section{Experiment}
\subsection{Experimental Setup}
We evaluate 21 advanced reasoning and non-reasoning LLMs from 7 model families, including Gemini 2.5 Pro/Flash, Deepseek-R1/V3, GPT-4o, Claude 3.7, Qwen-max/2.5/2, LLaMA 3.1/3, Gemma 3/2, on our MixRea benchmark. We access all models using APIs from Alibaba Cloud, OpenRouter, and OpenAI. All models are alignment (-chat or -instruct) models. We aim to investigate the following three questions:
\begin{enumerate}

\item Do LLMs exhibit inattentional blindness when solving the explicit-implicit reasoning task?
\item How can prompting mitigate inattentional blindness and enhance LLMs' explicit-implicit reasoning capabilities?
\item Does inattentional blindness in LLMs manifest in broader scenarios involving multi-source information integration and reasoning?

\end{enumerate}



\subsection{Experimental Results}
\paragraph{(Q1) Evaluating Inattentional Blindness in Explicit-Implicit Reasoning.} 
Figure~\ref{fig:main_result} presents an overview of model performance on MixRea, covering both accuracy and consistency metrics. Additional results across reasoning types can be found in Appendix D.

\paragraph{Accuracy.} 
Overall, model accuracy ranges from 0.27 to 0.68, highlighting that explicit-implicit reasoning remains a challenging task for current LLMs. Among the models evaluated, Gemini 2.5 Pro (67.9\%) and Deepseek-R1 (64.6\%) stand out, representing the best-performing closed-source and open-source models, respectively. In the non-reasoning category, LLaMA 3.1-405B (62.4\%) and Claude 3.7 Sonnet (62.1\%) achieve the highest accuracy, while the remaining models fall below the 60\% threshold.

\paragraph{Consistency.} 
Consistency measures whether a model produces coherent reasoning outcomes across different combinations of explicit and implicit information within the same initial context. All models exhibit relatively low consistency scores (below 43\%), with a typical gap exceeding 25\% between accuracy and consistency. This discrepancy suggests that LLMs often vary their reasoning behavior depending on the specific information presented, indicating a lack of stability in handling explicit-implicit inference. 

\paragraph{Error Analysis.}\label{sec:4.2.2 error analysis}

\begin{table}[t]
    \centering
    \begin{tabular}{lccc}
        \toprule
        \textbf{Model}  & \textbf{IIA}($\uparrow$) & \textbf{EIA}($\uparrow$) & \textbf{I3R}($\downarrow$)\\
        \midrule
        Gemini 2.5 Pro & \textbf{76.5} & 91.4 & \textbf{9.4} \\
        Deepseek-R1 & 70.9 & 92.4 & 9.5 \\
        Gemini 2.5 Flash & 67.1 & \textbf{92.5} & 15.1 \\
        Claude 3.7 Sonnet & 70.1 & 92.1 & 12.7 \\
        GPT-4o      & 66.6 & 91.5 & 13.2 \\
        GPT-4o-mini & 57.5 & 86.6 & 20.3 \\
        Deepseek-V3 & 65.4 & 92.5 & 14.0 \\
        Qwen-Max    & 65.0 & 91.6 & 15.4 \\
        Qwen 2.5-72B & 63.9 & 87.9 & 16.0 \\
        Qwen 2.5-32B & 66.4 & 91.7 & 14.6 \\
        Qwen 2.5-7B  & 62.6 & 84.1 & 15.0 \\
        Qwen 2-72B   & 61.3 & 91.6 & 17.7 \\
        Qwen 2-7B    & 64.2 & 85.1 & 14.3 \\
        LLaMA 3.1-405B& 71.1 & 91.1 & 12.4 \\
        LLaMA 3.1-70B& 67.6 & 90.5 & 13.6 \\
        LLaMA 3.1-8B & 54.9 & 83.3 & 20.3 \\
        LLaMA 3-70B  & 65.4 & 90.3 & 14.6 \\
        LLaMA 3-8B   & 57.5 & 81.7 & 18.3 \\
        Gemma 3-27B & 65.2 & 87.6 & 16.2 \\
        Gemma 3-12B & 61.8 & 86.6 & 18.2 \\
        Gemma 2-27B & 65.1 & 87.2 & 14.3 \\
        \bottomrule
    \end{tabular}
    \caption{The error analysis results based on three metrics: implicit information accuracy (IIA), explicit information accuracy (EIA), and implicit information ignorance rate (I3R).}
    \label{tab:comparison}
\end{table}

As shown in Table \ref{tab:comparison}, we provide an error analysis based on model explanations and three metrics:  explicit information accuracy (EIA), implicit information accuracy (IIA), and implicit information ignorance Rate (I3R). EIA and IIA represent the proportion of correctly reasoning with explicit and implicit information, respectively. I3R quantifies the proportion of cases where the model disregards implicit information.


\paragraph{Qualitative Analysis of Inattentional Blindness.} Explicit reasoning accuracy is generally higher than implicit reasoning accuracy, indicating the limitation lies in implicit reasoning. 
To determine whether implicit reasoning errors stem from the inherent difficulty of the task or from inattentional blindness caused by explicit information, we manually analyze the model's explanations for incorrect reasoning instances. Our analysis reveals that the primary source of errors is the model's tendency to focus exclusively on explicit question features while neglecting differences among candidate event sets in the options, thereby failing to recognize and reason about implicit information. These findings support the inattentional blindness hypothesis we aimed to investigate. More error examples containing model explanations are provided in Appendix E.

\paragraph{Quantitative Analysis of Inattentional Blindness.} Based on our option setting, the model selection process inherently incorporates corresponding reasoning mechanisms. We observed that all models exhibit a relatively high implicit information ignorance rate (9\%–21\%), with even the best-performing reasoning model, Gemini 2.5 Pro, reaching 9.4\%. This demonstrates that, even when excluding reasoning errors, the direct omission of implicit information due to inattentional blindness remains a major bottleneck in explicit-implicit reasoning for LLMs.


\paragraph{(Q2) Mitigating Inattentional Blindness through Prompting}
\begin{figure}[t]
  \includegraphics[width=\columnwidth]{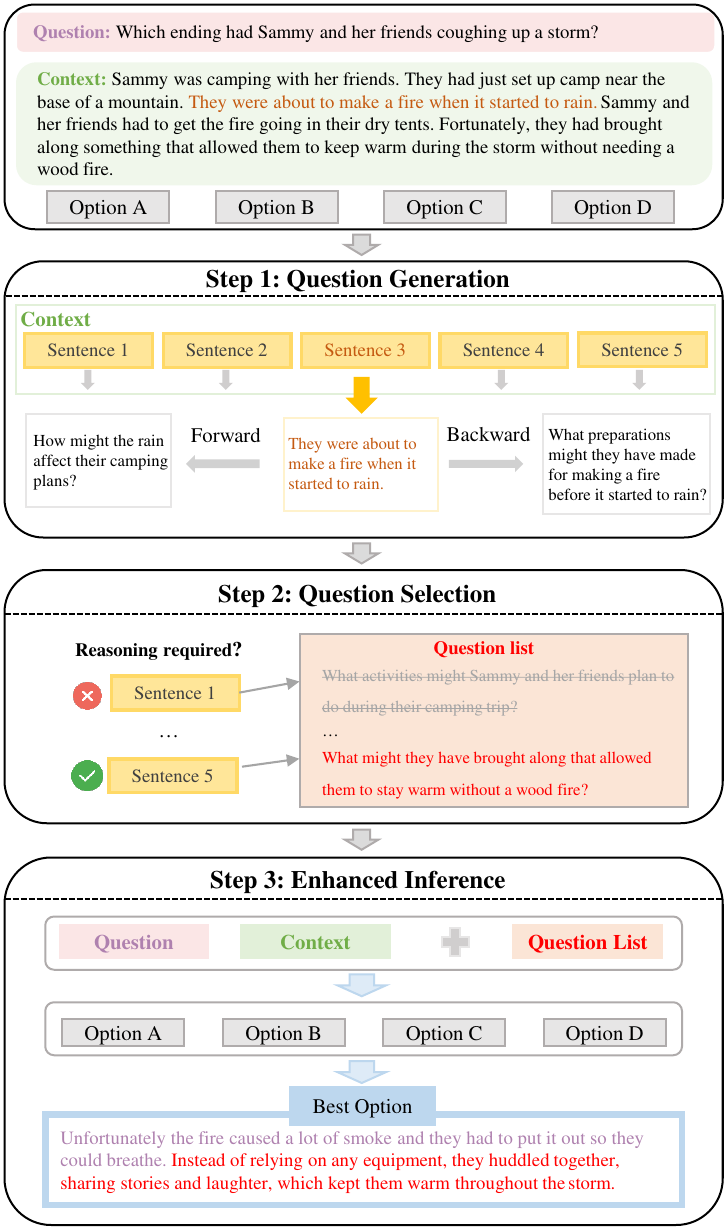}
  \caption{The illustration of our proposed PRCP prompting method.}
  \label{fig:prcp}
\end{figure}



In previous research, prompt-based approaches have been widely used to enhance the reasoning capabilities of LLMs \cite{yu-etal-2023-exploring-Legal-Reasoning-Tasks,zhou2023leasttomostpromptingenablescomplex}. As shown in Table \ref{tab:prompting_results}, classical prompt-based methods are difficult to achieve effective and stable improvements on MixRea. Therefore, we aim to explore how to mitigate inattentional blindness and enhance explicit-implicit reasoning abilities through prompt engineering.

\paragraph{Motivation.} Reasoning task instructions usually do not explicitly outline the relationships among task components, such as the context, question, and answer options. This requires LLMs to actively infer these connections. However, as shown in our error analysis, LLMs frequently miss implicit information during reasoning, revealing their difficulty in identifying latent associations between options and context. To address this limitation, we introduce the Potential Relationship Completion Prompting (PRCP) method, which strengthens explicit-implicit reasoning by recovering overlooked latent causal relationships.

\begin{table}[t]
    \centering
    \renewcommand{\arraystretch}{1} 
    \setlength{\tabcolsep}{1mm}
    \begin{tabular}{lcccc}
        \toprule
        \multirow{2}{*}{\textbf{Model}} & \multirow{2}{*}{\textbf{CoT}} & \multirow{2}{*}{\textbf{One-shot}} & \multicolumn{2}{c}{\textbf{PRCP}} \\
        \cmidrule(lr){4-5}
        & & & \textbf{w/o QS} & \textbf{w/ QS} \\
        \midrule
        Gemini 2.5 Flash & +1.3 & +1.2 & \textbf{+2.1} & +1.6 \\
        GPT-4o & +0.4 & +2.5 & +1.9 & \textbf{+3.0} \\
        GPT-4o-mini & +0.4 & 0.0 & +2.8 & \textbf{+3.7} \\
        Deepseek-V3 & \textbf{+2.6} & -3.1 & +0.7 & +1.6 \\
        Qwen-Max & +1.0 & +2.6 & \textbf{+4.0} & +3.6 \\
        LLaMA 3.1-405B & -1.8 & \textbf{+2.0} & +0.9 & +0.9 \\
        Qwen 2.5-72B & +0.6 & +0.8 & +3.9 & \textbf{+4.1} \\
        LLaMA 3.1-70B & -0.3 & +0.1 & +3.3 & \textbf{+3.4} \\
        Qwen 2.5-7B & -0.6 & -4.9 & +1.5 & \textbf{+2.4} \\
        LLaMA 3.1-8B & -0.8 & +1.7 & +3.7 & \textbf{+5.3} \\
        \bottomrule
    \end{tabular}
    \caption{Performance of LLMs across different prompting settings on MixRea. QS represents the question selection stage of PRCP.}
    \label{tab:prompting_results}
\end{table}

\begin{figure*}[ht]
  \includegraphics[width=\linewidth]{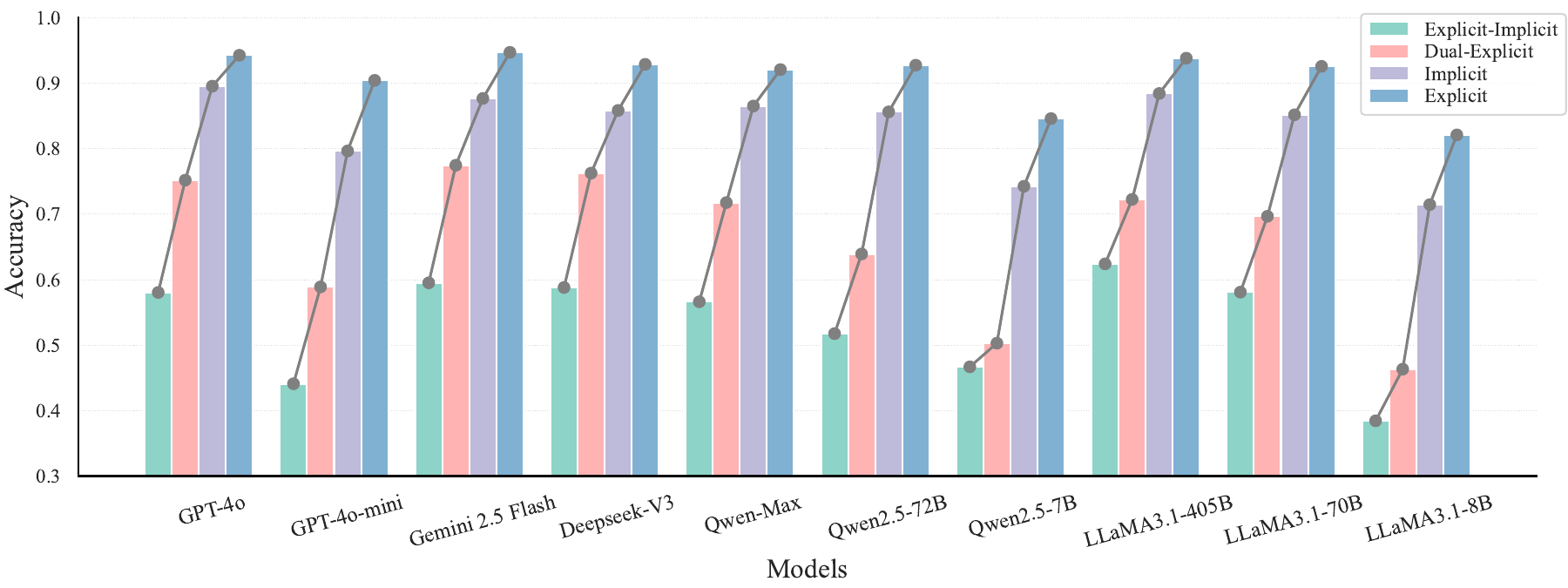}
  \caption{The comparison of results across four task settings: explicit-implicit reasoning, dual-explicit reasoning, implicit reasoning, and explicit reasoning.}
  \label{fig:four_setting_result}
\end{figure*}

\paragraph{Potential Relationship Completion Prompting.} As shown in Figure \ref{fig:prcp}, PRCP attempts to guide the LLMs' attention toward sentences in the context that may require reasoning by designing a chain of thoughts, thereby improving its ability to capture both explicit causal chains and implicit associations. Specifically, PRCP accomplishes these objectives through three collaborative stages: question generation, question selection, and enhanced inference.

\textbf{Step 1: Question Generation.} As a preliminary enhancement method, we explicitly prompt the model to generate additional forward and backward reasoning questions for each sentence in the text, creating a rich set of intermediate questions.  This active questioning mechanism forces the model to develop a deeper understanding of causal relationships and event progression within the text, helping it fill logical gaps in the narrative. Consequently, this strengthens LLMs' ability to identify and generate reasoning paths.

\textbf{Step 2: Question Selection.} In order to filter the generated intermediate reasoning questions, LLMs analyze the context sentence-by-sentence, identifying segments that require additional logical explanation or subsequent inference. The set of intermediate questions of the question generation stage is screened based on the identification results. This process aims to reduce noise from irrelevant questions, enhancing the quality and effectiveness of the question set.

\textbf{Step 3: Enhanced Inference.} The filtered intermediate reasoning questions are incorporated as supplementary information into the original reasoning task to enhance LLMs' explicit-implicit reasoning ability.


\paragraph{Performance and Analysis.}As shown in Table \ref{tab:prompting_results}, we compare the relative performance of CoT, one-shot, and PRCP method against the baseline without any prompting techniques. PRCP demonstrates consistent and significant performance improvements across all models, confirming its superior effectiveness and robustness compared to other methods. 
These results verify our hypothesis that LLMs' limited awareness of latent associations between task components is one contributing factor to inattentional blindness.

\paragraph{(Q3) Manifestation of Inattentional Blindness in Broader Multi-Source Reasoning}

To investigate whether inattentional blindness in LLMs extends to broader scenarios involving multi-source information integration and reasoning, we conducted additional experiments on MixRea using three extended task settings beyond explicit-implicit reasoning, performing ablation studies on the reasoning process. Furthermore, we explore the extent to which inattentional blindness impairs the reasoning capabilities of LLMs in different tasks. 
The other three tasks include dual-explicit reasoning, explicit reasoning, and implicit reasoning. Dual-explicit reasoning builds upon explicit-implicit reasoning by explicitly stating implicit information. Explicit reasoning and implicit reasoning focus on single-source information.

\paragraph{Inattentional Blindness in Boarder Scenario.} In Figure \ref{fig:four_setting_result}, we observe that for explicit reasoning and implicit reasoning, all large models maintain relatively high accuracy. This indicates that LLMs possess strong reasoning capabilities for both explicit and implicit information individually. However, accuracy significantly declines in dual-explicit reasoning and explicit-implicit reasoning compared to single-source reasoning. This suggests that LLMs exhibit general inattentional blindness in broader multi-source Information integration scenarios besides explicit-implicit reasoning.

\paragraph{Inattentional Blindness Influence in Different Tasks.} LLMs perform significantly worse in explicit-implicit reasoning tasks compared to dual-explicit reasoning, with a larger performance gap than that between explicit and implicit reasoning alone. This suggests that in multi-source reasoning, inattentional blindness is more pronounced when integrating explicit and implicit information than when processing multiple explicit information.

\section{Conclusion}

Explicit–implicit reasoning remains an under-investigated dimension of LLM cognition. Our study aims to fill this gap by introducing MixRea, a carefully curated benchmark to investigate whether current LLMs suffer from a form of inattentional blindness. A systematic evaluation of current advanced reasoning and non-reasoning models confirms a persistent failure to notice and integrate latent causal cues when tasked with explicit instructions. Guided by an in-depth error analysis, we further devise PRCP, a lightweight prompting strategy to mitigate the inattentional blindness and enhance explicit-implicit reasoning. PRCP consistently outperforms other methods by steering models to recover overlooked relations and to weave them into coherent reasoning chains. Complementary ablation studies on the reasoning process through four reasoning tasks demonstrate that varying degrees of inattentional blindness persist across more tasks involving multi-source information reasoning, indicating the significant impact of the limitation in LLMs.

\section{Acknowledgments}
This work was supported in part by the Science, Technology and Innovation Project of Shenzhen Longhua District (No. 20260309G23410662).

\bibliography{main}

\makeatletter
\@ifundefined{isChecklistMainFile}{
  \newif\ifreproStandalone
  \reproStandalonetrue
}{
  \newif\ifreproStandalone
  \reproStandalonefalse
}
\makeatother

\ifreproStandalone
\documentclass[letterpaper]{article}
\usepackage[submission]{aaai2026}
\setlength{\pdfpagewidth}{8.5in}
\setlength{\pdfpageheight}{11in}
\usepackage{times}
\usepackage{helvet}
\usepackage{courier}
\usepackage{xcolor}
\frenchspacing

\begin{document}
\fi
\setlength{\leftmargini}{20pt}
\makeatletter\def\@listi{\leftmargin\leftmargini \topsep .5em \parsep .5em \itemsep .5em}
\def\@listii{\leftmargin\leftmarginii \labelwidth\leftmarginii \advance\labelwidth-\labelsep \topsep .4em \parsep .4em \itemsep .4em}
\def\@listiii{\leftmargin\leftmarginiii \labelwidth\leftmarginiii \advance\labelwidth-\labelsep \topsep .4em \parsep .4em \itemsep .4em}\makeatother

\setcounter{secnumdepth}{0}
\renewcommand\thesubsection{\arabic{subsection}}
\renewcommand\labelenumi{\thesubsection.\arabic{enumi}}

\newcounter{checksubsection}
\newcounter{checkitem}[checksubsection]

\newcommand{\checksubsection}[1]{%
  \refstepcounter{checksubsection}%
  \paragraph{\arabic{checksubsection}. #1}%
  \setcounter{checkitem}{0}%
}

\newcommand{\checkitem}{%
  \refstepcounter{checkitem}%
  \item[\arabic{checksubsection}.\arabic{checkitem}.]%
}
\newcommand{\question}[2]{\normalcolor\checkitem #1 #2 \color{blue}}
\newcommand{\ifyespoints}[1]{\makebox[0pt][l]{\hspace{-15pt}\normalcolor #1}}

\section*{Reproducibility Checklist}

\vspace{1em}
\hrule
\vspace{1em}

\textbf{Instructions for Authors:}

This document outlines key aspects for assessing reproducibility. Please provide your input by editing this \texttt{.tex} file directly.

For each question (that applies), replace the ``Type your response here'' text with your answer.

\vspace{1em}
\noindent
\textbf{Example:} If a question appears as
\begin{center}
\noindent
\begin{minipage}{.9\linewidth}
\ttfamily\raggedright
\string\question \{Proofs of all novel claims are included\} \{(yes/partial/no)\} \\
Type your response here
\end{minipage}
\end{center}
you would change it to:
\begin{center}
\noindent
\begin{minipage}{.9\linewidth}
\ttfamily\raggedright
\string\question \{Proofs of all novel claims are included\} \{(yes/partial/no)\} \\
yes
\end{minipage}
\end{center}
Please make sure to:
\begin{itemize}\setlength{\itemsep}{.1em}
\item Replace ONLY the ``Type your response here'' text and nothing else.
\item Use one of the options listed for that question (e.g., \textbf{yes}, \textbf{no}, \textbf{partial}, or \textbf{NA}).
\item \textbf{Not} modify any other part of the \texttt{\string\question} command or any other lines in this document.\\
\end{itemize}

You can \texttt{\string\input} this .tex file right before \texttt{\string\end\{document\}} of your main file or compile it as a stand-alone document. Check the instructions on your conference's website to see if you will be asked to provide this checklist with your paper or separately.

\vspace{1em}
\hrule
\vspace{1em}


\checksubsection{General Paper Structure}
\begin{itemize}

\question{Includes a conceptual outline and/or pseudocode description of AI methods introduced}{(yes/partial/no/NA)}
yes

\question{Clearly delineates statements that are opinions, hypothesis, and speculation from objective facts and results}{(yes/no)}
yes

\question{Provides well-marked pedagogical references for less-familiar readers to gain background necessary to replicate the paper}{(yes/no)}
yes

\end{itemize}
\checksubsection{Theoretical Contributions}
\begin{itemize}

\question{Does this paper make theoretical contributions?}{(yes/no)}
yes

	\ifyespoints{\vspace{1.2em}If yes, please address the following points:}
        \begin{itemize}
	
	\question{All assumptions and restrictions are stated clearly and formally}{(yes/partial/no)}
	yes

	\question{All novel claims are stated formally (e.g., in theorem statements)}{(yes/partial/no)}
	yes

	\question{Proofs of all novel claims are included}{(yes/partial/no)}
	yes

	\question{Proof sketches or intuitions are given for complex and/or novel results}{(yes/partial/no)}
	yes

	\question{Appropriate citations to theoretical tools used are given}{(yes/partial/no)}
	yes

	\question{All theoretical claims are demonstrated empirically to hold}{(yes/partial/no/NA)}
	yes

	\question{All experimental code used to eliminate or disprove claims is included}{(yes/no/NA)}
	yes
	
	\end{itemize}
\end{itemize}

\checksubsection{Dataset Usage}
\begin{itemize}

\question{Does this paper rely on one or more datasets?}{(yes/no)}
yes

\ifyespoints{If yes, please address the following points:}
\begin{itemize}

	\question{A motivation is given for why the experiments are conducted on the selected datasets}{(yes/partial/no/NA)}
	yes

	\question{All novel datasets introduced in this paper are included in a data appendix}{(yes/partial/no/NA)}
	yes

	\question{All novel datasets introduced in this paper will be made publicly available upon publication of the paper with a license that allows free usage for research purposes}{(yes/partial/no/NA)}
	yes

	\question{All datasets drawn from the existing literature (potentially including authors' own previously published work) are accompanied by appropriate citations}{(yes/no/NA)}
	yes

	\question{All datasets drawn from the existing literature (potentially including authors' own previously published work) are publicly available}{(yes/partial/no/NA)}
	yes

	\question{All datasets that are not publicly available are described in detail, with explanation why publicly available alternatives are not scientifically satisficing}{(yes/partial/no/NA)}
	yes

\end{itemize}
\end{itemize}

\checksubsection{Computational Experiments}
\begin{itemize}

\question{Does this paper include computational experiments?}{(yes/no)}
yes

\ifyespoints{If yes, please address the following points:}
\begin{itemize}

	\question{This paper states the number and range of values tried per (hyper-) parameter during development of the paper, along with the criterion used for selecting the final parameter setting}{(yes/partial/no/NA)}
	yes

	\question{Any code required for pre-processing data is included in the appendix}{(yes/partial/no)}
	yes

	\question{All source code required for conducting and analyzing the experiments is included in a code appendix}{(yes/partial/no)}
	yes

	\question{All source code required for conducting and analyzing the experiments will be made publicly available upon publication of the paper with a license that allows free usage for research purposes}{(yes/partial/no)}
	yes
        
	\question{All source code implementing new methods have comments detailing the implementation, with references to the paper where each step comes from}{(yes/partial/no)}
	yes

	\question{If an algorithm depends on randomness, then the method used for setting seeds is described in a way sufficient to allow replication of results}{(yes/partial/no/NA)}
	yes

	\question{This paper specifies the computing infrastructure used for running experiments (hardware and software), including GPU/CPU models; amount of memory; operating system; names and versions of relevant software libraries and frameworks}{(yes/partial/no)}
	yes

	\question{This paper formally describes evaluation metrics used and explains the motivation for choosing these metrics}{(yes/partial/no)}
	yes

	\question{This paper states the number of algorithm runs used to compute each reported result}{(yes/no)}
	yes

	\question{Analysis of experiments goes beyond single-dimensional summaries of performance (e.g., average; median) to include measures of variation, confidence, or other distributional information}{(yes/no)}
	yes

	\question{The significance of any improvement or decrease in performance is judged using appropriate statistical tests (e.g., Wilcoxon signed-rank)}{(yes/partial/no)}
	yes

	\question{This paper lists all final (hyper-)parameters used for each model/algorithm in the paper’s experiments}{(yes/partial/no/NA)}
	yes

\end{itemize}
\end{itemize}
\ifreproStandalone
\end{document}
\fi

\end{document}